\pdfoutput=1
\documentclass[11pt]{article}

\usepackage[preprint]{acl}
\usepackage{flushend}

\usepackage{times}
\usepackage{latexsym}
\usepackage{amsmath}

\usepackage[T1]{fontenc}

\usepackage{pdfpages}
\usepackage{multirow}
\usepackage{amssymb}
\usepackage{booktabs}
\usepackage{pifont}
\usepackage{enumitem}
\usepackage{algorithm}
\usepackage{hyperref}
\usepackage{algpseudocode}
\usepackage{amsmath}
\usepackage{semantic}
\usepackage{color}
\NewDocumentCommand\emojismile{}{
    \includegraphics[scale=0.05]{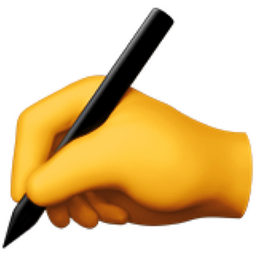}
}
\usepackage[utf8]{inputenc}

\usepackage{microtype}

\usepackage{inconsolata}

\usepackage[utf8]{inputenc}
\usepackage{graphicx}

\newcommand{\xmark}{\ding{55}}%

\title{\textsc{Commentator} \emojismile: A Code-mixed Multilingual Text Annotation Framework}

\author{
 \textbf{Rajvee Sheth\textsuperscript{$\dagger$}},
 \textbf{Shubh Nisar\textsuperscript{$\star$}},
 \textbf{Heenaben Prajapati\textsuperscript{$\dagger$}}, 
\\
 \textbf{Himanshu Beniwal\textsuperscript{$\dagger$}},
 \textbf{Mayank Singh\textsuperscript{$\dagger$}},
\\
\\
 \textsuperscript{$\dagger$}Indian Institute of Technology Gandhinagar,
 \textsuperscript{$\star$}North Carolina State University
\\
 \small{
   \textbf{Correspondence:} \href{mailto:lingo@iitgn.ac.in}{lingo@iitgn.ac.in}
 }
}

\begin{document}

\maketitle
\begin{abstract}
As the NLP community increasingly addresses challenges associated with multilingualism, robust annotation tools are essential to handle multilingual datasets efficiently. In this paper, we introduce a \textbf{\underline{co}}de-\textbf{\underline{m}}ixed \textbf{\underline{m}}ultilingual t\textbf{\underline{e}}xt a\textbf{\underline{n}}no\textbf{\underline{tat}}ion framew\textbf{\underline{or}}k, \textsc{Commentator}, specifically designed for annotating code-mixed text. The tool demonstrates its effectiveness in token-level and sentence-level language annotation tasks for \textbf{Hinglish} text. We perform robust qualitative human-based evaluations to showcase \textsc{Commentator} led to \textbf{5x} faster annotations than the best baseline. Our code is publicly available at \url{https://github.com/lingo-iitgn/commentator}. The demonstration video is available at \url{https://bit.ly/commentator_video}.

\end{abstract}

\section{Introduction}
Code mixing is prevalent in informal conversations and in social media, where elements from different languages are interwoven within a single sentence. A representative example in Hinglish such as ``\textit{I am feeling very thand today, so I'll wear a sweater.}''  (In this sentence, ``\textit{thand}'' is a Hindi word meaning ``\textit{cold}'', while the rest of the sentence is in English), demonstrating seamless integration of Hindi and English. A major challenge in NLP research is the scarcity of high-quality datasets, which require extensive manual efforts, significant time, domain expertise, and linguistic understanding, as highlighted by \citet{hovy2010towards}. The rise of social media has further complicated annotation tasks due to non-standard grammar, platform-specific tokens, and neologisms \citep{shahi2022amused}. Annotating these datasets presents unique challenges, including ensuring data consistency, efficiently managing large datasets,  mitigating annotator biases, and reporting poor-quality instances. Existing annotation tools often fail to address these diverse issues effectively.

\begin{figure}
    \centering
    \includegraphics[width=1\linewidth]{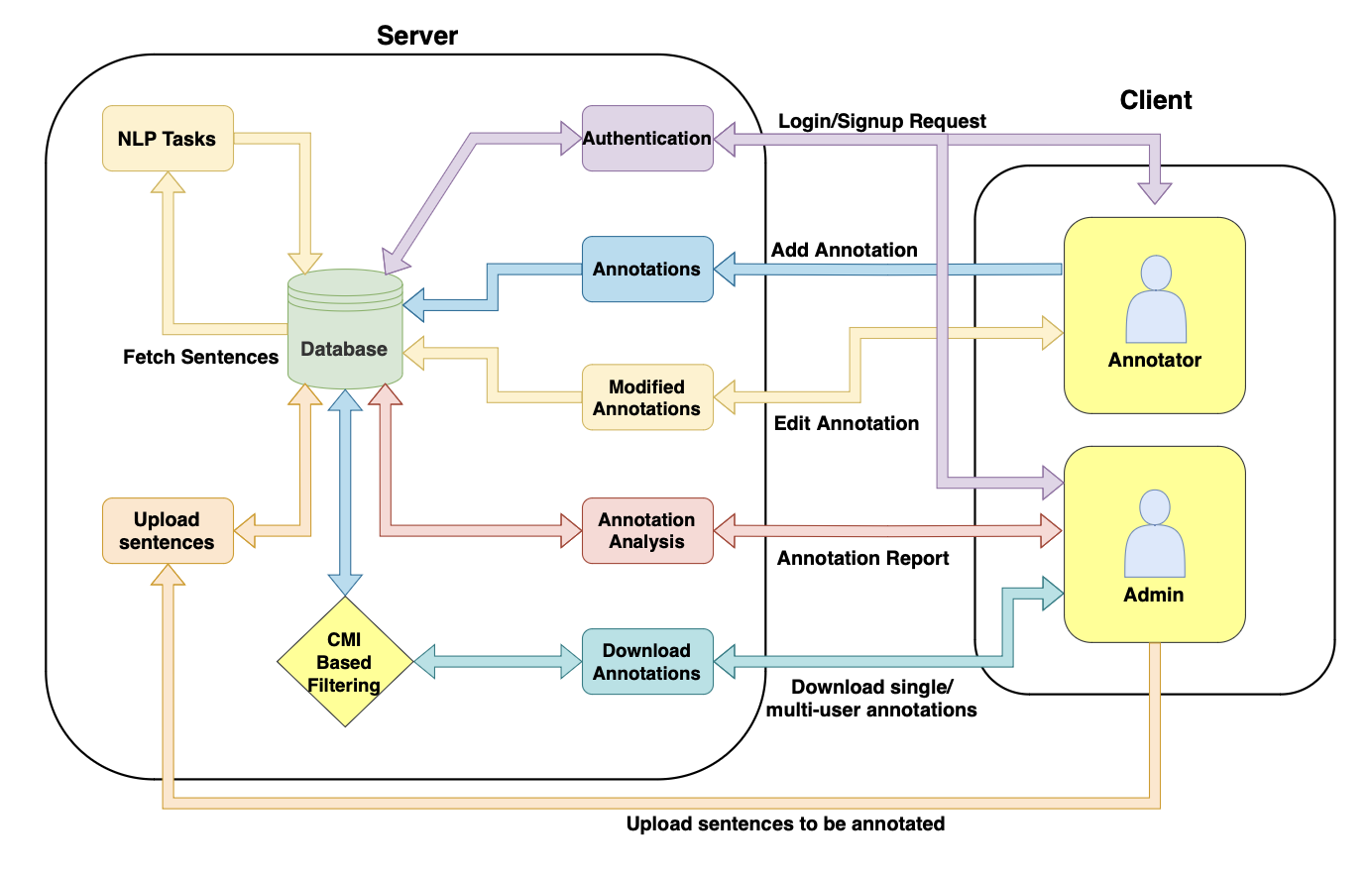}
    \caption{\textsc{Commentator} Framework.}
    \label{fig:architecture}
\end{figure}

This paper introduces \textsc{Commentator}, a robust annotation framework designed for multiple code-mixed annotation tasks. The current version\footnote{As a continual development effort, it will be further extended to three more popular code-mixing tasks NER, Spell Correction and Normalization, and Machine Translation.} of \textsc{Commentator} supports two token-level annotation tasks, \textbf{Language Identification}, \textbf{POS tagging}, and sentence-level \textbf{Matrix Language Identification}. While \textsc{Commentator} has already been used to generate a large number of annotations (more than 100K) in our ongoing project\footnote{URL available on our Github.}, these are not part of the current demo paper. The focus of this paper is to present the capabilities and initial functionalities of the framework. Figure~\ref{fig:architecture} presents the framework \textsc{Commentator}.

We evaluate \textsc{Commentator} by comparing its features and performance against five state-of-the-art text annotation tools, (i) YEDDA~\cite{yang2018yedda}, (ii)  MarkUp~\cite{dobbie2021markup}, (iii) INCEpTION~\cite{klie-etal-2018-inception}, (iv) UBIAI\footnote{\url{https://ubiai.tools/}} and (v) GATE~\cite{cunningham1996gate}. 
The major perceived capabilities (see Section~\ref{sec:time-comp-setup}) of \textsc{Commentator} are (i) simplicity in navigation and performing basic actions, (ii) task-specific recommendations to improve user productivity and ease the annotation process, (iii) quick cloud or local setup with minimal dependency requirements, (iv) promoting iterative refinement and quality control by integrating annotator feedback, (v) simple admin interface for uploading data, monitoring progress and post-annotation data analysis, and (vi) parallel annotations enabling multiple users to work on the same project simultaneously. Furthermore, Section~\ref{sec:time-comp-anno} demonstrates an annotation speed increase of nearly 5x compared to the nearest SOTA baseline. This speed gain can be further enhanced by incorporating more advanced code-mixed libraries.

In addition, the codebase, the demo website with a detailed installation guide, and some Hinglish sample instances are available on GitHub\footnote{\url{https://github.com/lingo-iitgn/commentator}}. 
Currently, the functionality is tailored for Hinglish, but it can be extended to support any language pair. 

\section{Existing Text Annotation Frameworks}
\label{sec:text_annotation}
Text annotation tools are vital in NLP for creating annotated datasets for training and evaluating machine learning models. This summary reviews several key tools, each with unique features and limitations.

\subsection{Web-based Annotation Tools}
These tools have been created to provide annotation environments independent of operating systems. Some of the web-based annotation tools are: \textbf{\textit{(1)}} \textit{MarkUp} improves annotation speed and accuracy using NLP and active learning but requires re-annotation for updates and has unreliable collaboration features \cite{dobbie2021markup}, \textbf{\textit{(2)}} \textit{INCEpTION} offers a versatile platform for semantic and interactive annotation but struggles with session timeouts and updating annotations \cite{klie-etal-2018-inception}, and lastly, \textbf{\textit{(3)}} \textit{UBIAI} provides advanced cloud-based NLP functions but faces problems with incorrect entity assignments and model integration \cite{ubiai2021}.

\subsection{Locally-hosted Tools} 
These tools can be installed on a local machine and offer more robust features or better performance for large datasets. Some of the locally hosted tools are: \textbf{\textit{(1)}} \textit{YEDDA} is an open source tool that enhances annotation efficiency and supports collaborative and administrative functions, though it has limitations in customization and can break tokens during annotation \cite{yang2018yedda}, \textbf{\textit{(2)}} \textit{GATE} is an open-source tool known for its real-time collaboration, but it is complicated to configure and slow with API requests \cite{bontcheva2013gate}, \textbf{\textit{(3)}} \textit{BRAT} is user-friendly for entity recognition and relationship annotation but lacks active learning and automatic suggestions \cite{stenetorp2012brat}, \textbf{\textit{(4)}} \textit{Prodigy} integrates with machine learning workflows and supports active learning but requires a commercial license \cite{montani2018prodigy}, and \textbf{\textit{(5)}} \textit{Doccano} is an open-source tool with a customizable interface for various annotation tasks but lacks advanced features like real-time collaboration \cite{nakayama2018doccano}.  Additional tools include \textbf{\textit{(6)}} \textit{Knowtator}, designed for biomedical annotations within \textit{Protégé}, but requires significant manual setup \cite{ogren-2006-knowtator}, \textbf{\textit{(7)}} \textit{WordFreak}, which is flexible but challenging for non-technical users \cite{morton-lacivita-2003-wordfreak}, \textbf{\textit{(8)}} \textit{Anafora}, known for its efficiency in biomedical annotation but lacking integration with machine learning models \cite{chen-styler-2013-anafora}, \textbf{\textit{(9)}} \textit{Atomic}, which is modular and powerful but requires extensive customization \cite{druskat2014atomic}, lastly, \textbf{\textit{(10)}} \textit{WebAnno} supports a wide range of annotation tasks and collaborative work, but encounters performance issues with large datasets \cite{yimam-etal-2013-webanno}.

While these tools offer diverse functionalities, each exhibits limitations that affect efficiency and usability. Most state-of-the-art frameworks are either paid or closed-source and do not support annotator feedback. Additionally, the majority do not enable parallel annotations over the internet and perform poorly when multiple scripts or words from different languages appear in the same sentence. The introduction of \textsc{Commentator} seeks to address these challenges by providing a robust framework specifically designed for multiple code-mixed annotation tasks.

\begin{figure}
\centering
\includegraphics[width=1\linewidth]{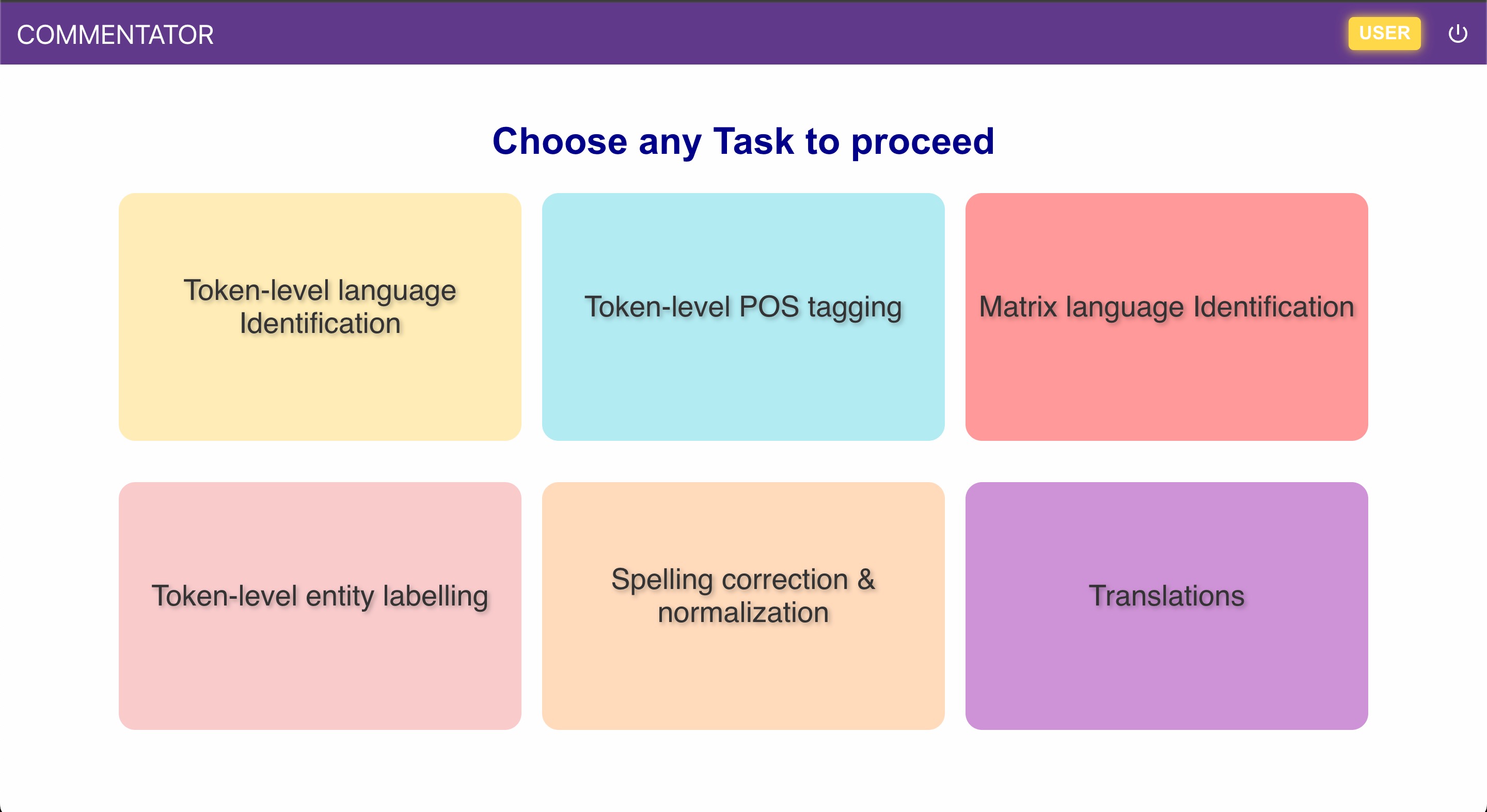} 
\caption{The Task interface of the \textsc{Commentator}.}\label{fig:task-interface}
\end{figure}

\begin{figure*}[!tbh]
\centering
\begin{tabular}{cc}
    \includegraphics[width=0.49\linewidth]{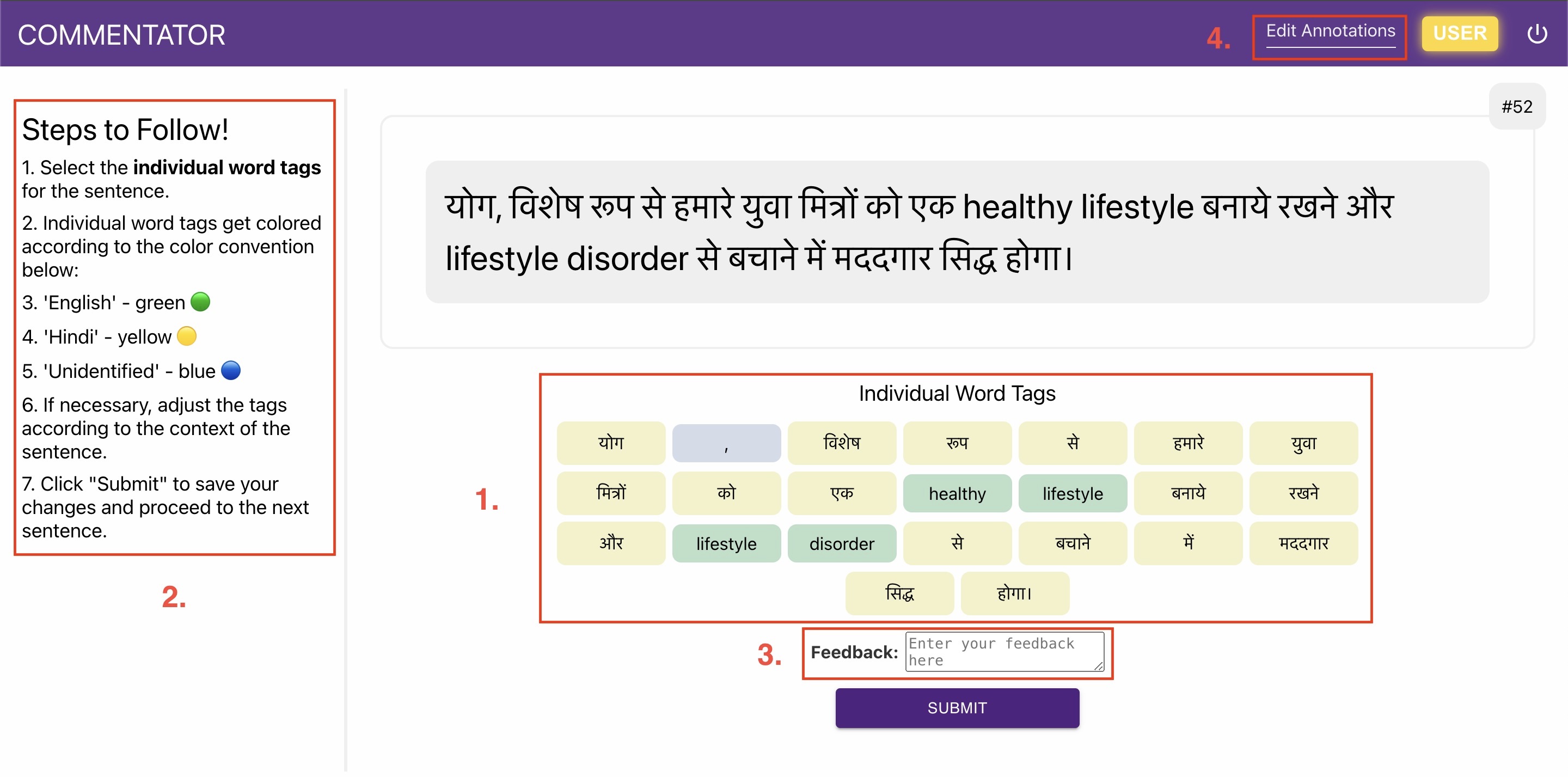} & 
    \raisebox{0\height}{\includegraphics[width=0.47\linewidth]{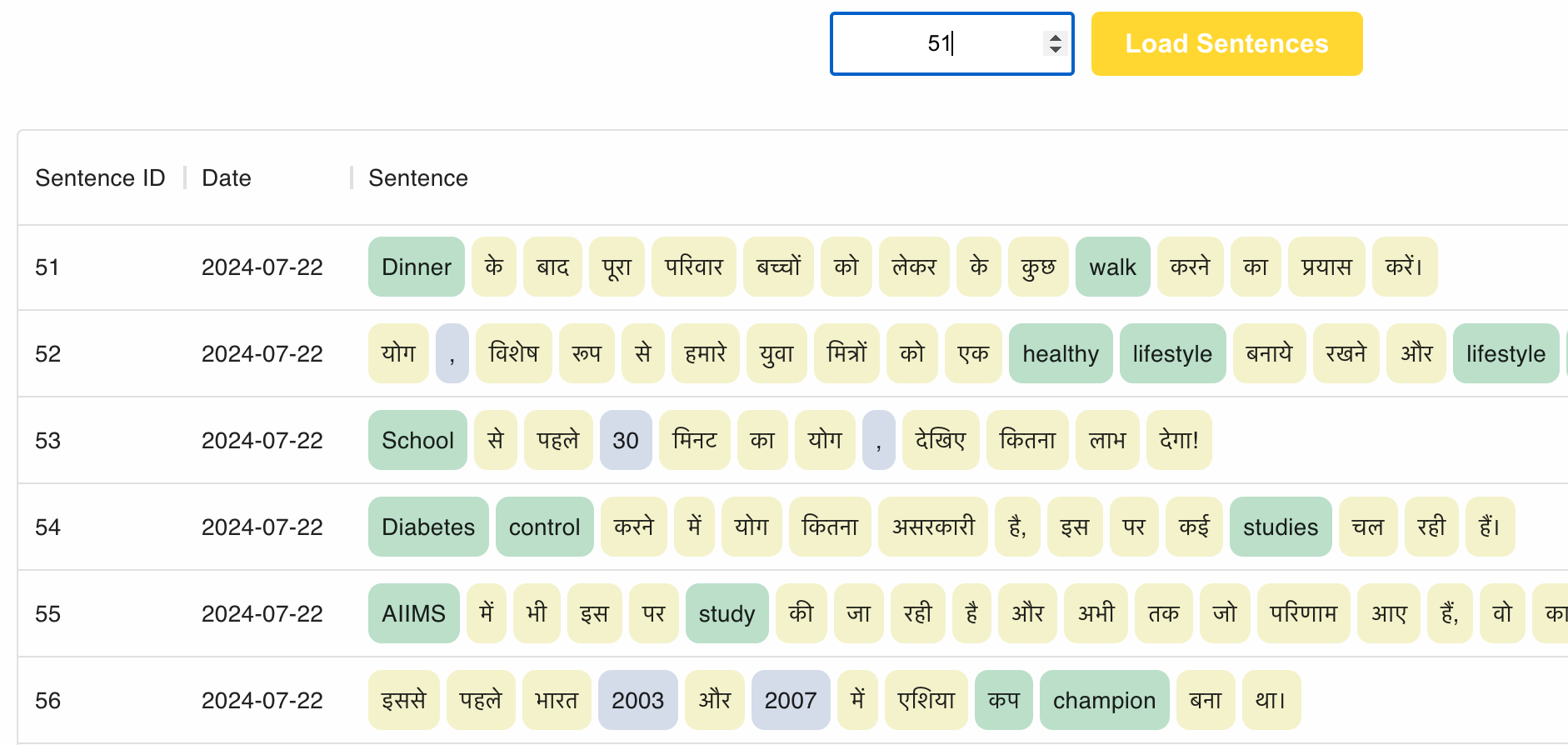}}\\
    (a) & (b) \\
\end{tabular}
\caption{Token-Level Language Identification (LID): (a) annotation page and (b) history and edit page.}\label{fig:annotator-interface}
\end{figure*}

\begin{figure*}
\centering
\begin{tabular}{cc}
    \includegraphics[width=0.49\linewidth]{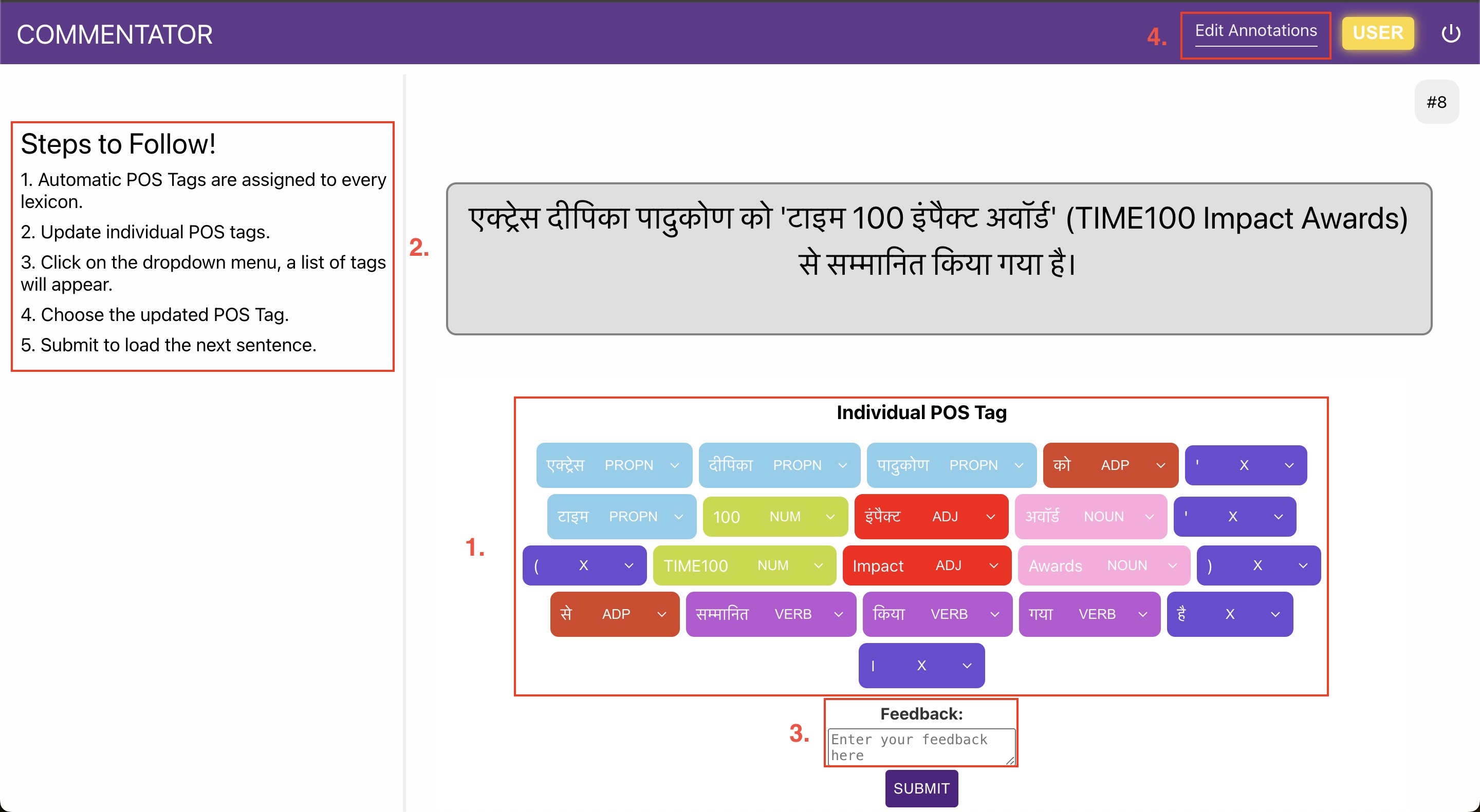} & 
    \raisebox{0.1\height}{\includegraphics[width=0.47\linewidth]{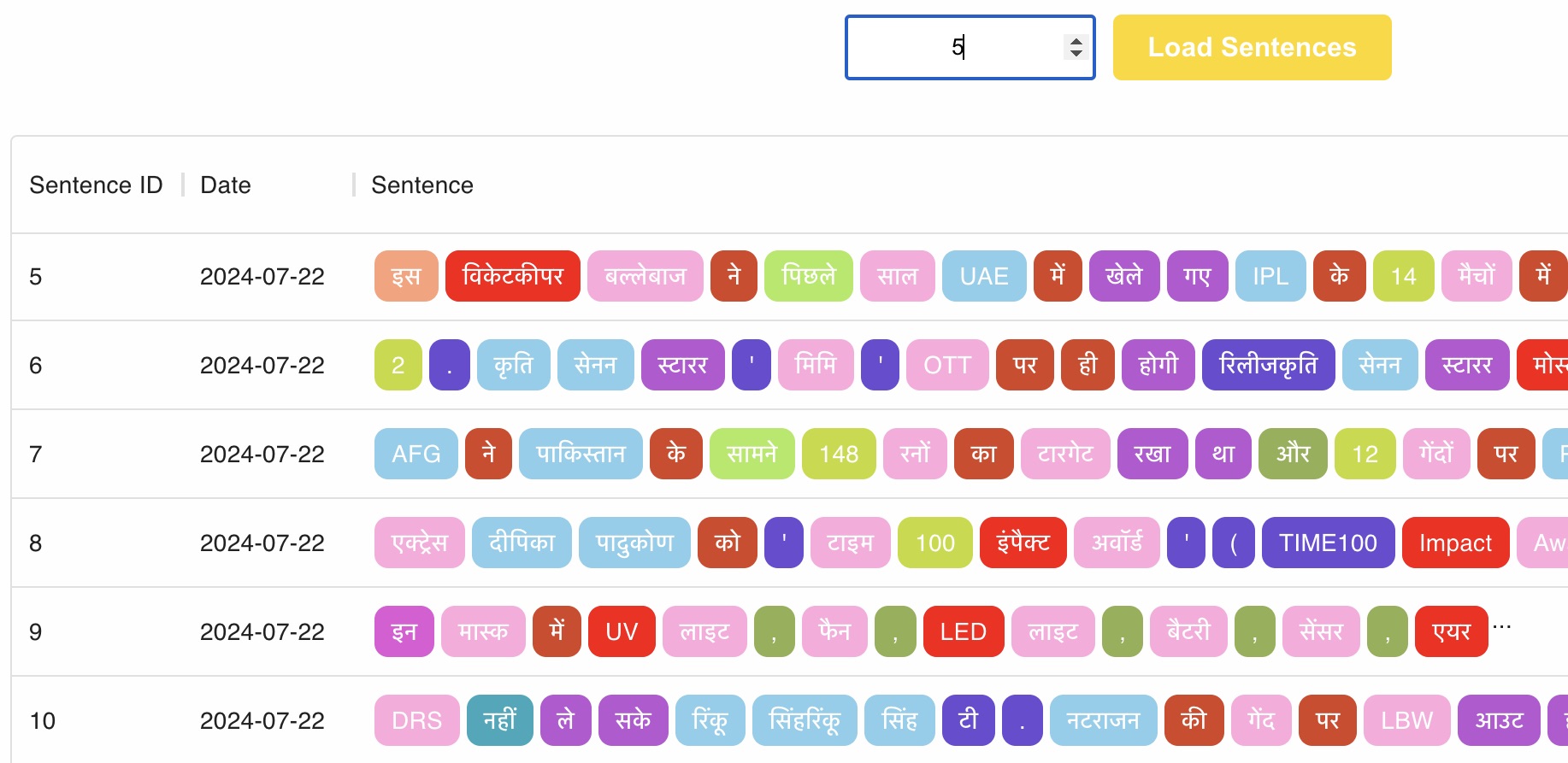}}\\ 
    (a)  & (b) \\
\end{tabular}
\caption{ Token-Level Parts-Of-Speech Tagging (POS): (a) annotation page and (b) history and edit page.}\label{fig:POS-interface}
\end{figure*}

\section{\textsc{COMMENTATOR}}

\subsection{The Functionalities}
\label{sec:commentator-function}
The proposed system caters to two types of users: \textbf{\textit{(i)}} the annotators and \textit{\textbf{(ii)}} the admins. Annotators perform annotation tasks. The admins design the annotation task, employ annotators, administer the annotation task, and process the annotations. Given these roles, we describe the \textsc{Commentator} functionalities by introducing:

\subsubsection{The Annotator Panel}
\label{sec:annotator_panel}
\begin{figure*}
\centering
\begin{tabular}{cc}
    \includegraphics[width=0.5\linewidth]{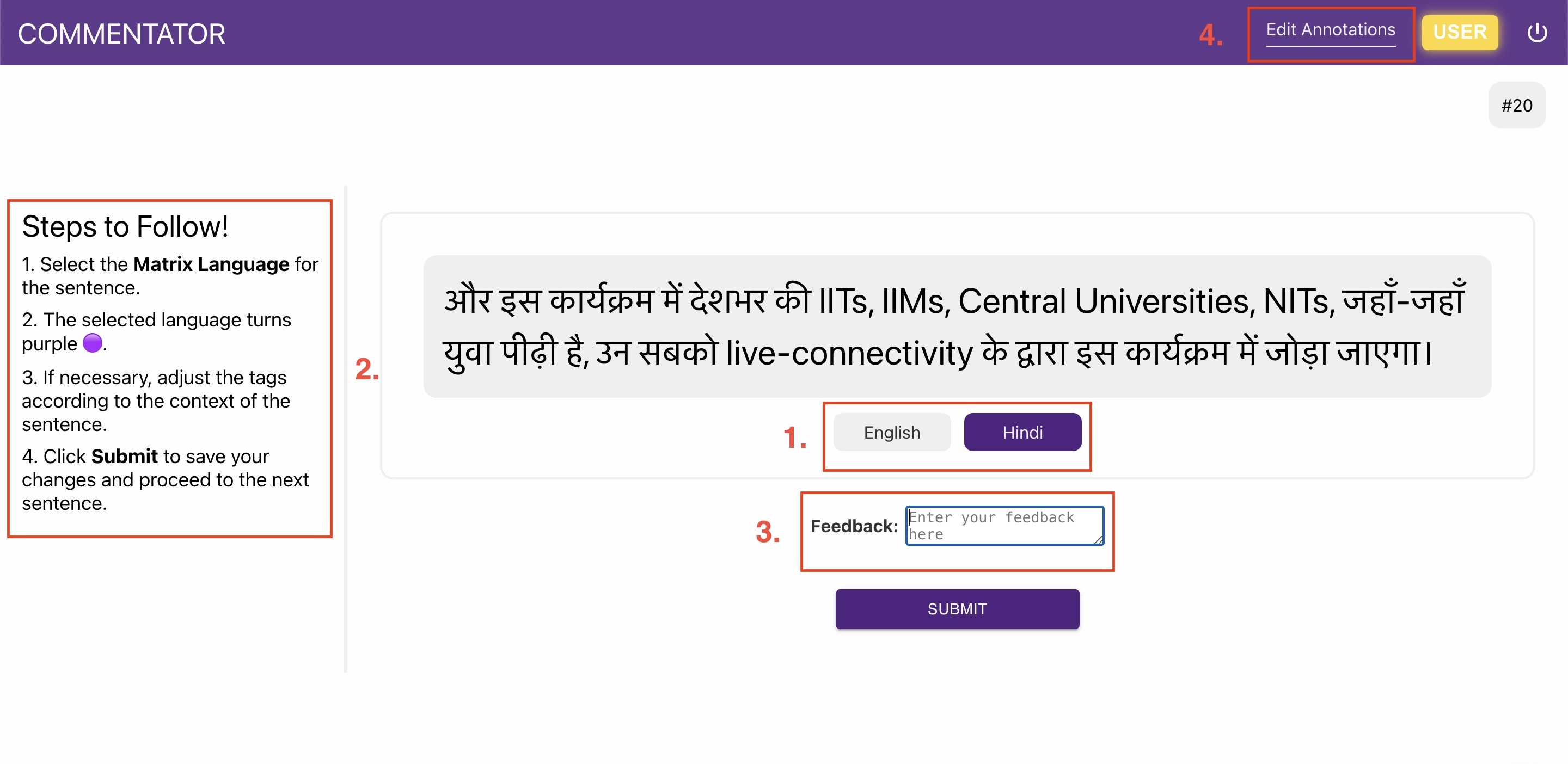} & 
    \raisebox{0.2\height}{\includegraphics[width=0.47\linewidth]{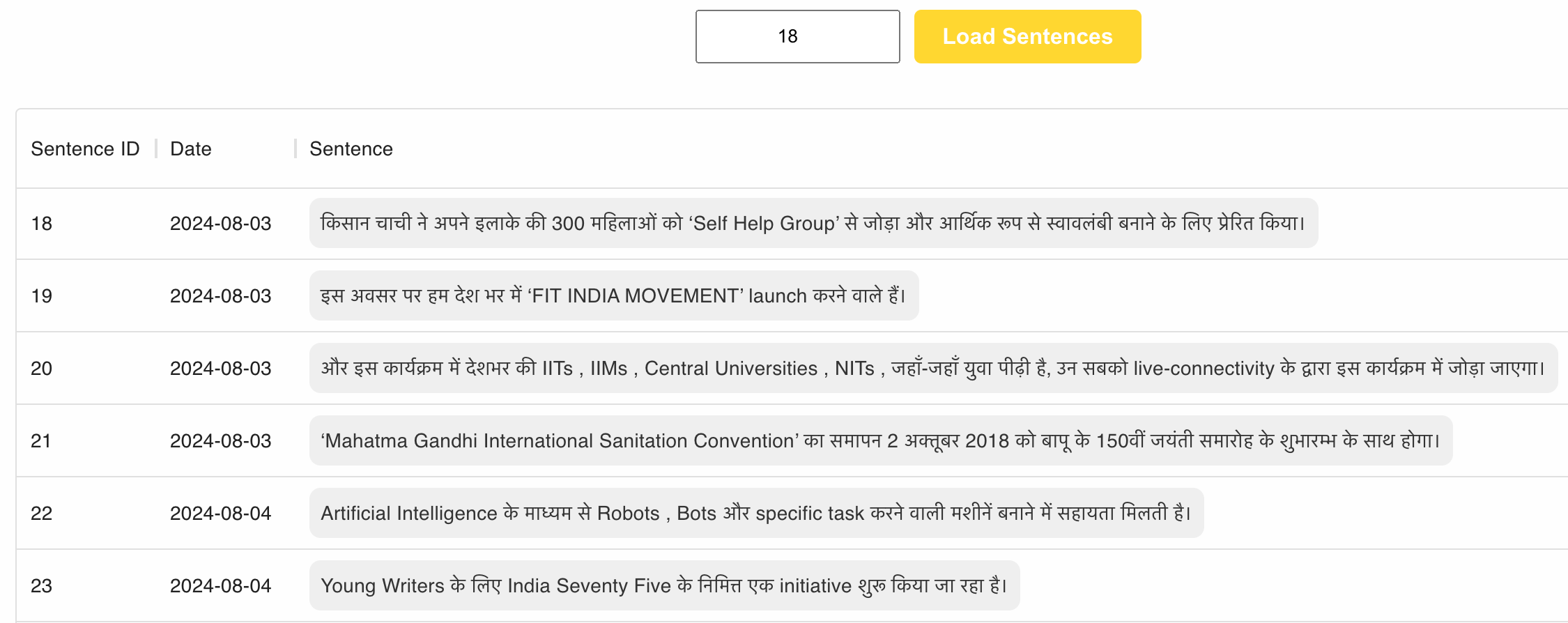}}\\ 
    (a)  & (b) \\
\end{tabular}
\caption{Matrix Language Identification (MID): (a) annotation page and (b) history and edit page.}\label{fig:Matrix-interface}
\end{figure*}

The annotator panel contains three pages: 
\begin{enumerate}[noitemsep,nolistsep,leftmargin=*]
    \item \textit{Landing page}: Figure~\ref{fig:task-interface} presents an annotator landing page. Here, the annotators are presented with a selection of several NLP tasks, displayed as clickable options. Selecting a task directs them to the dedicated annotation page for that specific task. 

    \item \textit{Annotation pages}: We, next, describe annotation pages for the first three tasks:
    \begin{itemize}
    \item \textbf{Token-Level Language Identification (LID):} 
    This task involves identifying the language of individual words (tokens) within a sentence (Figure~\ref{fig:annotator-interface}a, point \textbf{1}). Each token is pre-assigned a language tag using a state-of-the-art language identification API \footnote{\url{https://github.com/microsoft/LID-tool}\label{fn:lid-tool}}(more details are presented in Section~\ref{sec:commentator-server}). Annotators can update these tags by clicking the tag button until the desired tag appears.  Textual feedback can be entered in the ``\textit{Enter Your Feedback Here}'' section (Figure~\ref{fig:annotator-interface}a, point \textbf{3}). Textual feedback is essential to highlight issues with the current sentence. Some issues include grammatically incorrect sentences, incomplete sentences, sensitive/private information, toxic content, etc. 

    \item \textbf{Token-Level Parts-Of-Speech Tagging (POS):} 
    Similar to LID, this task involves identifying the POS tags of individual tokens within a text. Each token is pre-assigned a language tag using a state-of-the-art POS tagging CodeSwitch NLP library \footnote{\url{https://github.com/sagorbrur/codeswitch}\label{fn:codeswitch}}(more details are presented in Section~\ref{sec:commentator-server}).  In case of incorrect assignment of the tag, the annotators can select the correct tag from a drop-down menu (Figure~\ref{fig:POS-interface}a, point \textbf{1}). We do not keep the toggling button feature due to many POS tags. Similarly to LID, annotators can provide feedback (Figure~\ref{fig:POS-interface}a, point \textbf{3}).

    \item \textbf{Matrix Language Identification (MLI):} As shown in Figure~\ref{fig:Matrix-interface}, this task involves identifying the language that provides the syntactic structure of a code-mixed sentence. Annotators select the matrix language from the multiple supported languages for each sentence (Figure~\ref{fig:Matrix-interface}, point \textbf{1}). 
    
    \end{itemize}
\noindent The primary instructions are present on the left side of the page for each task (See point \textbf{2} in Figures~\ref{fig:annotator-interface}a,~\ref{fig:POS-interface}a and~\ref{fig:Matrix-interface}a). Similarly, annotations can be corrected by clicking the ``Edit Annotations'' button (see point \textbf{4} in Figures~\ref{fig:annotator-interface}a,~\ref{fig:POS-interface}a and~\ref{fig:Matrix-interface}a), which redirects to the corresponsing \textit{history and edit} pages (see Figures~\ref{fig:annotator-interface}b,~\ref{fig:POS-interface}b and~\ref{fig:Matrix-interface}b).

    \item \textit{History and Edit pages}: Figures~\ref{fig:annotator-interface}b, ~\ref{fig:POS-interface}b and~\ref{fig:Matrix-interface}b show a list of previously annotated sentences with timestamps for LID, POS and MLI, respectively. Clicking on a sentence opens the respective annotation page with the previously chosen tags for editing.
    
    \end{enumerate}

\subsubsection{The Admin Panel}
\label{sec:admin_panel}
Figure~\ref{fig:admin-interface} shows the admin panel. The admin panel performs three major tasks:

\begin{enumerate}[noitemsep,nolistsep,leftmargin=*]
    \item \textit{Data upload}: The administrator can upload the source sentences using a CSV file (Figure~\ref{fig:admin-interface}, point \textbf{1}).
   
    \item \textit{Annotation analysis}: The administrator can: \textit{\textbf{(i)}} analyze the quality of annotations using Cohen's Kappa score for inter-annotator agreement (IAA) (Figure~\ref{fig:admin-interface}, point \textbf{3}) and \textbf{\textit{(ii)}} analyze the degree of code-mixing in the annotated text using the code-mixing index (CMI) \cite{das-gamback-2014}\footnote{The CMI score ranges from 0 (monolingual) to 100 (highly code-mixed).}(Figure~\ref{fig:admin-interface}, point \textbf{2}).
    
    \item \textit{Data download}: The admin can \textit{download} annotations of single/multiple annotators in a CSV file. Admins can select specific tasks from a dropdown menu to customize the data extraction (Figure~\ref{fig:admin-interface}, point \textbf{2}) The data download functionality also supports the conditional filtering of data based on \textit{IAA} and \textit{CMI}.
    
\begin{figure}
\centering
\begin{tabular}{c}
    \includegraphics[width=\linewidth]{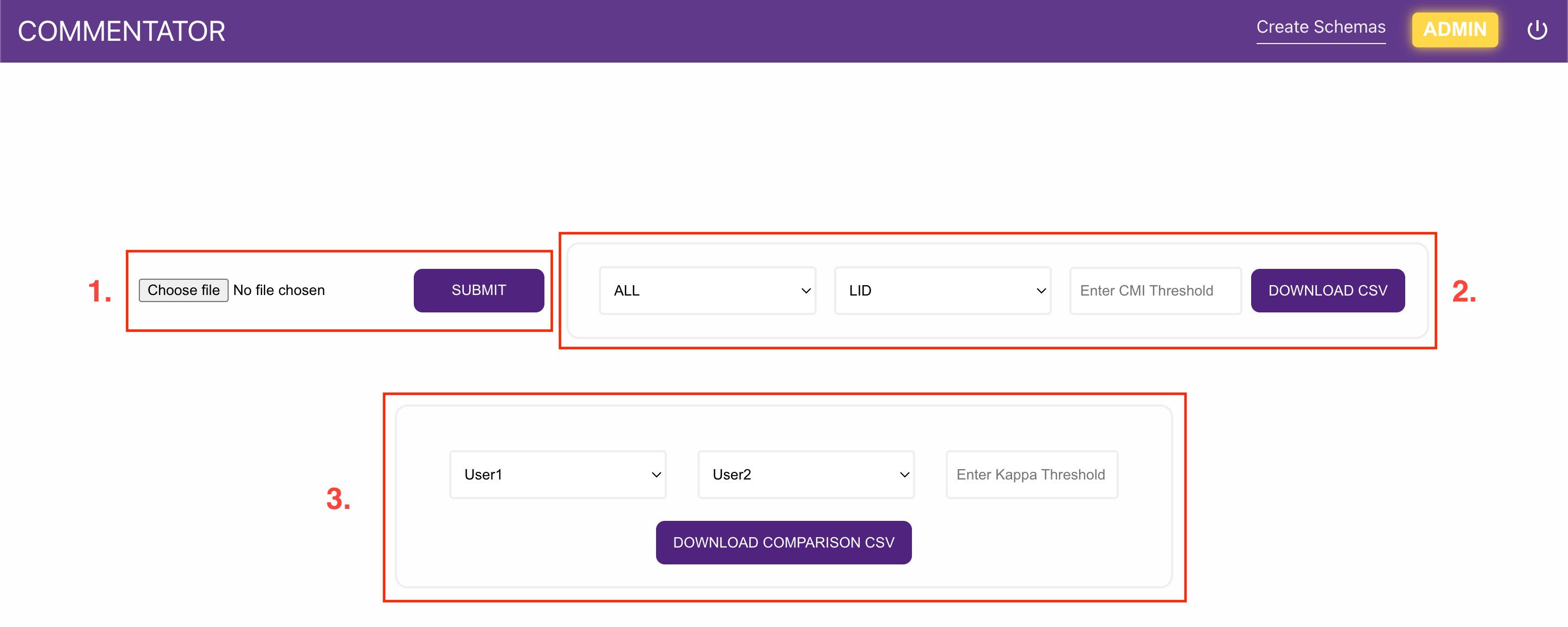}\\
\end{tabular}
\caption{The admin interface of the \textsc{Commentator}.}
\label{fig:admin-interface}
\end{figure}

\end{enumerate}

\begin{table*}
    \centering
    \resizebox{\hsize}{!}{
    \begin{tabular}{l|ccc|ccc|ccc|ccc|ccc|ccc}
      \hline
      \multirow{2}{*}{\textit{Capabilities}} 
      & \multicolumn{3}{|c|}{\textit{YEDDA}}
      & \multicolumn{3}{|c|}{\textit{MarkUp}}
      & \multicolumn{3}{|c|}{\textit{INCEpTION}} 
      & \multicolumn{3}{|c|}{\textit{UBIAI}}
      & \multicolumn{3}{|c|}{\textit{GATE}}
      & \multicolumn{3}{|c}{\textit{\textsc{Commentator}}}\\
      & 1 & 2 & 3 & 1 & 2 & 3 & 1 & 2 & 3  & 1 & 2 & 3 & 1 & 2 & 3  & 1 & 2 & 3  \\ \hline
     Operational ease &  \textcolor{red}{\xmark} &  \textcolor{red}{\xmark} & \textcolor{green}{\checkmark} &  \textcolor{green}{\checkmark} &  \textcolor{green}{\checkmark} &  \textcolor{red}{\xmark} & \textcolor{green}{\checkmark} &  \textcolor{red}{\xmark} & \textcolor{red}{\xmark} &  \textcolor{red}{\xmark} & \textcolor{green}{\checkmark} & \textcolor{green}{\checkmark} &  \textcolor{red}{\xmark} & \textcolor{red}{\xmark} & \textcolor{red}{\xmark} & \textcolor{green}{\checkmark} &  \textcolor{green}{\checkmark} & \textcolor{green}{\checkmark}\\

      Less dependency requirements & \textcolor{green}{\checkmark} & \textcolor{green}{\checkmark} & \textcolor{green}{\checkmark} & \textcolor{green}{\checkmark} & \textcolor{green}{\checkmark} & \textcolor{green}{\checkmark} & \textcolor{red}{\xmark} & \textcolor{red}{\xmark} &\textcolor{green}{\checkmark} &\textcolor{red}{\xmark} & \textcolor{green}{\checkmark} & \textcolor{green}{\checkmark} & \textcolor{red}{\xmark} & \textcolor{green}{\checkmark} & \textcolor{green}{\checkmark} & \textcolor{green}{\checkmark} & \textcolor{green}{\checkmark} & \textcolor{green}{\checkmark} \\

    Low latency in API requests  & \textcolor{red}{\xmark} & \textcolor{red}{\xmark} & \textcolor{red}{\xmark} & \textcolor{red}{\xmark} & \textcolor{green}{\checkmark} & \textcolor{red}{\xmark} & \textcolor{red}{\xmark} & \textcolor{red}{\xmark} & \textcolor{green}{\checkmark} &\textcolor{green}{\checkmark} & \textcolor{red}{\xmark} & \textcolor{red}{\xmark} & \textcolor{green}{\checkmark} & \textcolor{red}{\xmark} & \textcolor{green}{\checkmark} & \textcolor{green}{\checkmark} & \textcolor{green}{\checkmark} & \textcolor{green}{\checkmark} \\\hline
    
        Admin Interface &  \textcolor{green}{\checkmark} & \textcolor{green}{\checkmark} & \textcolor{green}{\checkmark} & \textcolor{green}{\checkmark} & \textcolor{green}{\checkmark} & \textcolor{green}{\checkmark} & \textcolor{green}{\checkmark} & \textcolor{green}{\checkmark} &\textcolor{green}{\checkmark} &\textcolor{green}{\checkmark} & \textcolor{green}{\checkmark} & \textcolor{green}{\checkmark} & \textcolor{green}{\checkmark} & \textcolor{green}{\checkmark} & \textcolor{green}{\checkmark} & \textcolor{green}{\checkmark} & \textcolor{green}{\checkmark} & \textcolor{green}{\checkmark} \\
      
      System recommendation & \textcolor{green}{\checkmark} & \textcolor{green}{\checkmark} & \textcolor{red}{\xmark} & \textcolor{red}{\xmark} & \textcolor{red}{\xmark} & \textcolor{red}{\xmark} & \textcolor{green}{\checkmark} & \textcolor{green}{\checkmark} & \textcolor{red}{\xmark} & \textcolor{green}{\checkmark} &  \textcolor{green}{\checkmark} & \textcolor{green}{\checkmark} & \textcolor{green}{\checkmark} & \textcolor{red}{\xmark} & \textcolor{red}{\xmark} & \textcolor{green}{\checkmark} & \textcolor{green}{\checkmark} & \textcolor{green}{\checkmark}\\

     Multiple user collaboration & \textcolor{red}{\xmark} & \textcolor{red}{\xmark} & \textcolor{red}{\xmark} & \textcolor{red}{\xmark} & \textcolor{green}{\checkmark} & \textcolor{red}{\xmark} & \textcolor{green}{\checkmark} & \textcolor{green}{\checkmark} &\textcolor{green}{\checkmark} &\textcolor{green}{\checkmark} & \textcolor{green}{\checkmark} & \textcolor{green}{\checkmark} & \textcolor{red}{\xmark} & \textcolor{red}{\xmark} & \textcolor{red}{\xmark} & \textcolor{green}{\checkmark} & \textcolor{green}{\checkmark} & \textcolor{green}{\checkmark} \\
     
      Annotation Refinement and Feedback & \textcolor{green}{\checkmark} & \textcolor{red}{\xmark} & \textcolor{red}{\xmark} & \textcolor{red}{\xmark} & \textcolor{green}{\checkmark} & \textcolor{green}{\checkmark} & \textcolor{green}{\checkmark} & \textcolor{red}{\xmark} &\textcolor{red}{\xmark} &\textcolor{green}{\checkmark} & \textcolor{green}{\checkmark} & \textcolor{green}{\checkmark} & \textcolor{green}{\checkmark} & \textcolor{red}{\xmark} & \textcolor{green}{\checkmark} & \textcolor{green}{\checkmark} & \textcolor{green}{\checkmark} & \textcolor{green}{\checkmark} \\
      
      Post-annotation analysis & \textcolor{green}{\checkmark} & \textcolor{green}{\checkmark} & \textcolor{green}{\checkmark} & \textcolor{red}{\xmark} & \textcolor{red}{\xmark} & \textcolor{red}{\xmark} & \textcolor{green}{\checkmark} & \textcolor{green}{\checkmark} &\textcolor{green}{\checkmark} &\textcolor{red}{\xmark} & \textcolor{green}{\checkmark} & \textcolor{red}{\xmark} & \textcolor{red}{\xmark} & \textcolor{red}{\xmark} & \textcolor{red}{\xmark} & \textcolor{green}{\checkmark} & \textcolor{green}{\checkmark} & \textcolor{green}{\checkmark}  \\ \hline
    \end{tabular}}
    \caption{Perceived capabilities by annotators. All annotators perceive all the eight capabilities in \textsc{Commentator}.}
    \label{tab:comp_cap}
\end{table*}

\subsection{The Architecture}
\label{sec:commentator-architecture}
Figure~\ref{fig:architecture} showcases the highly modular architecture for \textsc{Commentator}. We describe it using two main modules:

\subsubsection{Client Module}
\label{sec:commentator-client}
The client is developed using \textit{ReactJS}\footnote{\url{https://reactjs.org}}. The client module comprises pages for the following functionalities: \textbf{\textit{(i)}} User Login, \textit{\textbf{(ii)}} User Signup, \textbf{\textit{(iii)}} Annotation Panel, and \textbf{\textit{(iv)}} History, and \textit{\textbf{(v)}} Admin Panel. The user login page is used to log into the portal. The user signup page creates a new annotator account on the portal. The annotation panel is the main landing page that initiates the annotation process for all tasks. The history page lists the annotated sentences by the logged-in annotator for individual tasks.

\subsubsection{Server Module}
\label{sec:commentator-server}
The client is served using a Flask\footnote{\url{https://flask.palletsprojects.com/en/2.1.x/}} Server. The server performs two major functions: \textbf{\textit{(i)}} connection with the database and \textbf{\textit{(ii)}} calling task-specific API/libraries. It connects to the \textit{MongoDB} database through a Pymongo library. The MongoDB database can be locally hosted or on the cloud. We use the MongoDB Atlas database\footnote{\url{https://www.mongodb.com/atlas/database}} hosted locally. In the current setup, we use Microsoft API for LID\footnote{Existing open source libraries such as Spacy-LangDetect (\url{https://pypi.org/project/spacy-langdetect/}) and LangDetect (\url{https://pypi.org/project/langdetect/}) showed poor performance}. For POS, we use the CodeSwitch NLP library. This also demonstrates the flexibility of \textsc{Commentator} to make web-based API calls or local-hosted library calls based on the task requirements. 

\section{Experiments}
\label{sec:exp}
In this section, we perform two human studies to evaluate \textit{\textsc{Commentator}} against recent state-of-the-art tools to ensure a comprehensive comparison with modern advancements and cutting-edge functionalities: (i) YEDDA~\cite{yang2018yedda}, (ii) MarkUp~\cite{dobbie2021markup}, (iii) INCEpTION~\cite{klie-etal-2018-inception}, (iv) UBIAI\footnote{\url{https://ubiai.tools/}}, and (v) GATE~\cite{bontcheva2013gate}. The first study assesses the total time and perceived capabilities during the initial low-level setup and at higher-level annotation tasks (see Section~\ref{sec:time-comp-setup} for more details). The second study examines the annotation time (see Section~\ref{sec:time-comp-anno} for more details).

\subsection{Initial Setup and Perceived Capabilities}
\label{sec:time-comp-setup}
We employ three human annotators proficient in English and Hindi with experience using social media platforms such as X (formally `Twitter'). Additionally, the annotators are graduate students with good programming skills and knowledge of version control systems. Each annotator has a detailed instruction document\footnote{\url{https://github.com/lingo-iitgn/commentator/tree/main/Documents}\label{fn:docs}}
containing links to execute codebases or access the web user interface, descriptions of tool configurations, annotation processes, and guidelines for recording time. 

Each annotator measures the time taken for the initial setup, including installation and configuration. The initial setup includes installation (downloading source code, decompressing, and installing dependencies) and configuration (adding configuration files, sentence loading, and user account creation/login).:

\begin{enumerate}[noitemsep,nolistsep,leftmargin=*]
\item \textit{Operational  Ease}: 
    A tool demonstrates operational ease when it requires minimal effort for installation, data input, and output. A user-friendly interface with features like color gradients for tag differentiation enhances the annotation experience, leading to more engaging and prolonged usage compared to tools with less visually appealing interfaces.
   
    \item \textit{Less Dependency Requirements}:
    Annotation tools often require resolving multiple dependencies during installation, which is challenging due to rapid advancements in web frameworks, data processing pipelines, and programming languages. This complexity limits usage, particularly among non-CS users.

    \item \textit{Low Latency in API Requests}:
    Latency is measured as the time to serve the request made by a client. This is the main bottleneck in web-based annotation tools that deal with APIs to serve and process data.
     
    \item \textit{Admin Interface}:
    The tool should feature an intuitive admin interface for efficient user management, role assignment, and annotation progress monitoring, offering comprehensive control without requiring extensive technical knowledge.

    \item \textit{System Recommendation}:
    Effective system recommendations that use advanced NLP tools and APIs can streamline the annotation process and reduce the annotation time.

    \item \textit{Parallel Annotations}:
    The tool should support multiple users to work simultaneously on the same dataset, share insights, and maintain consistency across annotations, enhancing overall efficiency and reliability.

    \item \textit{Annotation Refinement and Feedback}:
    The tool must allow annotators to refine and update their annotations easily.

    \item \textit{Post-annotation Analysis}: This feature evaluates annotation quality using metrics like inter-annotator agreement, with statistical measures like Cohen's Kappa (it gauges the degree of consistency among annotations), enhancing the reliability and validity of the data. In addition, as the 
    \textsc{Commentator} largely focuses on the code-mixed domain; integration of metrics like Code-mixing Index (CMI) is highly preferred.
    
\end{enumerate}

\noindent Annotators report each tool's setup time and assign a ``Yes/No'' label to eight perceived capabilities. Table~\ref{tab:comp_time_installation} reports the time taken in seconds for five baselines tool and \textsc{Commentator}. Overall, YEDDA takes the least time to install and configure. However, Table~\ref{tab:comp_cap} presents a slightly more distinct picture. \textsc{Commentator} receives all eight perceived capabilities, while all existing state-of-the-art annotation frameworks, except UIBAI, lack operational ease. Additionally, none of the tools possess a feedback mechanism that allows users to report any inconsistencies during annotations, including identifying noisy or abusive datasets for potential removal. All annotators agree that YEDDA exhibits poor user collaboration capabilities. 

\begin{table}[t]
    \centering
    \resizebox{\hsize}{!}{
    \begin{tabular}{lcc}
      \toprule
      \textit{Tools} & \textit{Installation}& \textit{Configuration} \\ \toprule
      YEDDA &\textbf{7.66$\pm$8.73}& \textbf{24.33$\pm$32.29} \\
      MarkUp &NA& 366.67$\pm$47.25\\
      INCEpTION  &NA& 247.66$\pm$39.80\\
      UBIAI  &NA& 324.33 $\pm$ 62.90 \\
     GATE  & 45.67$\pm$11.44 & 125.00 $\pm$ 68.07 \\\hline
     \textsc{Commentator} (ours) & 173.33$\pm$89.93 & 210.00$\pm$81.65 \\  \bottomrule
    \end{tabular}}
    \caption{Comparison of time taken (mean $\pm$ standard deviation) for installation and configuration in seconds. `NA' corresponds to those web-based tools that cannot be installed on local systems. YEDDA takes the least time to install and configure.  \textsc{Commentator}'s configuration time is lower than three popular tools, MarkUp, INCEpTION and UBIAI.}
    \label{tab:comp_time_installation}
\end{table}

\begin{table}[t]
    \centering
    \resizebox{\hsize}{!}{
    \begin{tabular}{lcc}
      \toprule
      \textit{Tools} & \textit{LID}& \textit{POS} \\ \toprule
      YEDDA &757.00$\pm$62.27& 1370.66$\pm$81.24 \\
      MarkUp &{1192.33$\pm$172.77}& 1579.00$\pm$68.86\\
      INCEpTION  &{1040.66$\pm$69.67}& 1714.66$\pm$71.30\\
      UBIAI  & 690.66$\pm$ 79.43 & 748.33$\pm$91.45\\
     GATE  & 1118.33$\pm$166.20 & 1579.00 $\pm$ 50.61 \\\hline
     \textsc{Commentator} (ours) & \textbf{138.33 $\pm$ 24.60} & \textbf{337.66 $\pm$ 25.34} \\  \bottomrule
    \end{tabular}}
    \caption{Comparison of time taken (mean $\pm$ standard deviation) for annotation in seconds. \textit{POS}, being a highly challenging task than LID, took significantly more time. LID annotations on \textsc{Commentator} are \textbf{5x} faster than the next best tool, UBIAI. Whereas POS annotations on \textsc{Commentator} are \textbf{2x} faster than UBIAI.}
    \label{tab:comp_time_annotation}
\end{table}

\subsection{Annotation Time}
\label{sec:time-comp-anno}
In the second human study, we recruit three annotators with a good understanding of Hindi and English languages\footnote{The three annotators recruited in the first human study are different than these annotators.}. Each annotator annotates ten Hinglish sentences (available on the project's GitHub page) for token-level language tasks: (i) LID and (ii) POS. Both tasks involve assigning a tag to each token in a sentence. For LID, the tags are \textit{Hindi}, \textit{English}, \textit{Unidentified}. For POS, we follow the list of tags proposed by \citet{singh-etal-2018-twitter}. This list includes \textit{NOUN}, \textit{PROPN}, \textit{VERB}, \textit{ADJ}, \textit{ADV}, \textit{ADP}, \textit{PRON}, \textit{DET}, \textit{CONJ}, \textit{PART}, \textit{PRON\_WH}, \textit{PART\_NEG}, \textit{NUM}, and \textit{X}. Here, X denotes foreign words, typos, and abbreviations. Table~\ref{tab:comp_time_annotation} shows that the libraries that preassign tags enable \textsc{Commentator} to perform at least five times faster in annotation than the existing tools. 

\noindent Overall, annotators find that \textsc{Commentator} takes slightly longer time in initial setup but significantly reduces annotation time and efforts. It showcases good recommendation capability, parallel annotations and post-annotation analysis capabilities.  

\section{Conclusion and Future Work}
\label{sec:conc}
We introduce \textsc{Commentator}, an annotation framework for code-mixed text, and compared it against five state-of-the-art annotation tools. \textsc{Commentator} shows better user collaboration, operational ease, and efficiency, significantly reducing annotation time for tasks like Language Identification and Part-of-Speech tagging. Future plans include expanding \textsc{Commentator} to support tasks such as sentiment analysis, Q\&A, and language generation, making it an even more comprehensive tool for multilingual and code-mixed text annotation.

\newpage
\section{Ethics}
We adhere to the ethical guidelines by ensuring the responsible development and use of our annotation tool. Our project prioritizes annotator well-being, data privacy, and bias mitigation while promoting transparency and inclusivity in NLP research.

\bibliography{custom}
\newpage
\appendix
\newpage

\section{Appendix}

\subsection{Inter-annotator agreement (IAA)}
\label{sec:kohen}
 IAA measures how well multiple annotators can make the same annotation decision for a particular category. IAA shows you how clear your annotation guidelines are, how uniformly your annotators understand them, and how reproducible the annotation task is. Cohen’s kappa coefficient \cite{Hallgrenandkevin, cohen1960coefficient} is a statistic to measure the reliability between annotators for qualitative (categorical) items. It is a more robust measure than simple percent agreement calculations, as k considers the possibility of the agreement occurring by chance. It is a pairwise reliability measure between two annotators.

The formula for Cohen's kappa (\(\kappa\)) is:
\begin{equation}
    \kappa = \frac{P_o - P_e}{1 - P_e}
\end{equation}

where, $P_o$ is \textit{relative observed agreement among raters} and $P_e$ is \textit{hypothetical probability of chance agreement}. 

\subsection{Code-mixing Index (CMI)}
\label{sec:CMI}

CMI metric~\cite{das2014identifying} is defined as follows:
\begin{equation}
\label{eq:oldCMI}
    CMI= \begin{cases} 
100 * [1- \frac{max(w_{i})}{n-u}] & n> u \\
0 & n=u 
\end{cases}
\end{equation} 

Here, $w_{i}$ is the number of words of the language $i$, max\{{$w_{i}$}\} represents the number of words of the most prominent language, $n$ is the total number of tokens, $u$ represents the number of language-independent tokens (such as named entities, abbreviations, mentions, and hashtags). A low CMI score indicates monolingualism in the text whereas the high CMI score indicates the high degree of code-mixing in the text.

\section{Limitations}

We present some of the limitations in the \textsc{Commentator} tool, along with potential areas for future improvement:
\begin{enumerate}
    \item \textbf{Web-hosting}: \textsc{Commentator} is not currently web-based, but we are developing a web version to improve accessibility and user experience.
    \item \textbf{Model Integration}: The tool does not yet support direct integration of pre-trained models through the user interface for predictions.
    \item \textbf{Post-annotation Analysis}: While offering basic post-annotation analysis, future versions will include task-specific metrics such as Fleiss' Kappa, Krippendorff's Alpha, and Intraclass Correlation for more detailed evaluations of inter-annotator reliability and annotation accuracy.
\end{enumerate}

\section{Acknowledgements}
This work is supported by the Science and Engineering Research Board (SERB) through the project titled ``Curating and Constructing Benchmarks and Development of ML Models for Low-Level NLP Tasks in Hindi-English Code-Mixing''. The authors express their gratitude to Diksha, Mahesh Kumar, and Ronakpuri Goswami for their invaluable support with annotation. We also extend our thanks to Vannsh Jani, Isha Narang, and Eshwar Dhande for their assistance in reviewing the manuscript and reporting on installation and configuration times.

\end{document}